\documentclass[10pt,letterpaper]{article}
\usepackage[margin=1.3 in]{geometry}
\usepackage{amsthm}

\usepackage{natbib}

\newcommand{\ba}{\boldsymbol{a}}

\newcommand{\be}{\boldsymbol{e}}

\usepackage{bm}

\newcommand{\bt}{\boldsymbol{t}}
\newcommand{\bu}{\boldsymbol{u}}
\newcommand{\bv}{\boldsymbol{v}}
\newcommand{\bw}{\boldsymbol{w}}
\newcommand{\bx}{\boldsymbol{x}}
\newcommand{\by}{\boldsymbol{y}}
\newcommand{\bz}{\boldsymbol{z}}

\newcommand{\bA}{\boldsymbol{A}}

\newcommand{\bD}{\boldsymbol{D}}

\newcommand{\bI}{\boldsymbol{I}}

\newcommand{\bK}{\boldsymbol{K}}

\newcommand{\bQ}{\boldsymbol{Q}}

\newcommand{\bU}{\boldsymbol{U}}
\newcommand{\bV}{\boldsymbol{V}}

\newcommand{\bZ}{\boldsymbol{Z}}
\newcommand{\balpha}{\boldsymbol{\alpha}}
\newcommand{\bbeta}{\boldsymbol{\beta}}

\newcommand{\bepsilon}{\boldsymbol{\epsilon}}

\newcommand{\bsigma}{\boldsymbol{\sigma}}

\newcommand{\bSigma}{\boldsymbol{\Sigma}}

\newcommand{\E}{\mathbb{E}}

\usepackage[british]{babel}

\usepackage{mathtools}
\DeclarePairedDelimiter{\norm}{\lVert}{\rVert}

\usepackage[utf8]{inputenc} 
\usepackage[T1]{fontenc}    
\usepackage{hyperref}       
\usepackage{url}            
\usepackage{booktabs}       
\usepackage{amsfonts}       
\usepackage{nicefrac}       
\usepackage{microtype}      
\usepackage{xcolor}         
\usepackage{amsmath}
\usepackage{amsfonts}
\usepackage{amssymb}

\usepackage{graphicx,subcaption,lipsum}
\usepackage{bm}
\usepackage{tabularx}
\usepackage{booktabs}
\usepackage{subcaption}

\usepackage[dvipsnames]{xcolor}
\usepackage{algorithm}
\usepackage{algorithmic}
\newtheorem{theorem}{Theorem}

\theoremstyle{plain}

\theoremstyle{definition}

\theoremstyle{remark}
\newtheorem{remark}{Remark}

\usepackage[verbose]{wrapfig}

\usepackage{hyperref}
\hypersetup{
    colorlinks=true,
    linkcolor=blue,
    filecolor=magenta,
    citecolor=BlueViolet,
    urlcolor=BlueViolet,
    pdftitle={Overleaf Example},
    pdfpagemode=FullScreen,
    }

\title{A New Framework for Convex Clustering in Kernel Spaces: Finite Sample
Bounds, Consistency and Performance Insights}

\author{
  Shubhayan Pan$^{1}$,
  Kushal Bose$^{2}$,
  Debolina Paul$^{3}$, \\
  Saptarshi Chakraborty$^{4}$,
  Swagatam Das$^{2}$ \\[0.8em]
  $^{1}$Indian Statistical Institute, Kolkata \\
  $^{2}$Electronics and Communication Sciences Unit, Indian Statistical Institute \\
  $^{3}$Department of Statistics, University of Oxford \\
  $^{4}$Department of Statistics, University of Michigan \\[0.6em]
  \text{shubhayanpan@gmail.com},
  \text{kushalbose92@gmail.com}, \\
  \text{debolina.paul@stats.ox.ac.uk},
  \text{saptarsc@umich.edu},\\
  \text{swagatam.das@isical.ac.in}
}

\begin{document}

\date{}

\maketitle

\begin{abstract}
Convex clustering is a well-regarded clustering method, resembling the
similar centroid-based approach of Lloyd's $k$-means, without requiring a
predefined cluster count. It starts with each data point as its centroid
and iteratively merges them. Despite its advantages, this method can fail
when dealing with data exhibiting linearly non-separable or non-convex
structures. To mitigate the limitations, we propose a kernelized extension
of the convex clustering method. This approach projects the data points
into a Reproducing Kernel Hilbert Space (RKHS) using a feature map,
enabling convex clustering in this transformed space. This kernelization
not only allows for better handling of complex data distributions but also
produces an embedding in a finite-dimensional vector space. We provide a
comprehensive theoretical underpinning for our kernelized approach,
proving algorithmic convergence and establishing finite sample bounds for
our estimates. The effectiveness of our method is demonstrated through
extensive experiments on both synthetic and real-world datasets, showing
superior performance compared to state-of-the-art clustering techniques.
This work marks a significant advancement in the field, offering an
effective solution for clustering in non-linear and non-convex data
scenarios. we release our code base at \href{https://github.com/Shubhayan29/Kernel-Convex-Clustering/tree/main}{https://github.com/Shubhayan29/Kernel-Convex-Clustering/tree/main}
\end{abstract}

\section{Introduction}
Convex clustering is one of the modern frameworks for performing a
clustering task, formulating it as a convex optimisation problem, thus
ensuring a unique and globally optimal solution. It leverages a fusion
penalty to enhance the grouping of the data, helping us to uncover hidden
structures in the data. It garnered widespread attention as an alternative
avenue that offers relaxations of traditionally non-convex problems
\citep{article}.  Given $n$ data points, $\bx_1, \dots, \bx_n \in \mathbb{R}^d$, convex clustering initially assumes $n$ distinct centroids
$\ba_1, \ldots, \ba_n \in \mathbb{R}^d$ for each of the $n$ points, and minimises the objective function given by
\begin{equation}
  \min_{\ba_1,\dots,\ba_n}\ \frac{1}{2}\sum_{i=1}^{n}\|\bx_i - \ba_i\|^2_2
  + \gamma \sum_{i < j} w_{ij}\|\ba_i - \ba_j\|_q
  \label{eq:convex_clust}
\end{equation}
Here $\|\cdot\|_q$ denotes the $\ell_q$ norm in $\mathbb{R}^d$, for some
$q \geq 1$. The first term measures the fit between $\bx_i$'s and $\ba_i$'s,
while the latter is a fusion term that penalizes the number of unique
$\ba_i$'s by way of an $\ell_q$ norm penalty with tuning parameter $\gamma$.
The weights $w_{ij}$ can be chosen heuristically to accelerate computation
and improve empirical performance. It is noteworthy that, for $q \geq 1$,
the objective is convex in $\ba_i$'s, and thus has a minimizer. This convex nature of the objective is attractive from a theoretical viewpoint: works by \cite{tan2015statisticalpropertiesconvexclustering}, \cite{62704d6c-b0ac-3303-9709-f518254d3b51} provide centroid recovery
guarantees, and \cite{chi2018recoveringtreesconvexclustering} establish conditions
under which the solution path recovers a tree. Apart from this, it has
many other attractive theoretical properties, that has garnered growing
interest in it
\citep{inproceedings,5967659,NIPS2014_3c9d14ca}.

In convex clustering, the number of clusters can be chosen automatically,
equating it to the number of distinct $\bu_i$'s. Indeed, the solution of
convex clustering offers a continuous path based on the parameter
$\gamma$, where a larger $\gamma$ increases the fusion penalty's
influence, leading to fewer unique centres or clusters \citep{Chi_2015}.

Over the years, different variants of convex clustering have been proposed
by different researchers. Some of the recent advances include SpaCC
\citep{Nagorski2016GenomicRD} for detecting genomic regions, ACC
\citep{chu2021adaptive} for convex clustering in generalized linear models,
TROUT \citep{9414417} for clustering of time series. Most of these variants
are data/application specific, reducing their general effectiveness. The
reader is advised to refer to \cite{10158737} for furthering their knowledge about the different
variants of convex clustering.

On the other hand, kernel methods emerge as a relevant preprocessing step
in clustering, as they can identify non-linear data patterns, which
conventional clustering techniques overlook. By employing the kernel
trick, kernel clustering methods map the data into a higher-dimensional
feature space, where clusters are linearly separable. Kernel $k$-means
\citep{6790375,1000150} extends the classical $k$-means algorithm by
incorporating kernel functions such as the Gaussian or polynomial kernels,
allowing the algorithm to identify complex, non-linear cluster boundaries
\citep{6790375}. This method has proven particularly effective in
applications like image segmentation and bioinformatics, where the data
often has several intricate structures that are not well identified by
linear methods \citep{1000150}. Kernel power $k$ means \citep{9928792} is
one of the many recent applications of kernel methods in the field of
clustering. Other applications in the clustering regime mostly include
multi-view clustering like \cite{Park2025SparseKKM,10529609,Wu_Feng_Yuan_2024,LI2024102086}.

Furthermore, \cite{NIPS2014_3c9d14ca} studied convex clustering from a
theoretical perspective, providing crucial details on a perfect cluster
recoveries and other related properties. Additionally, they tried to
kernelize convex clustering and formulated it as a second-order cone
optimization problem, but did not mention any details regarding its
implementation or any other theoretical analyses.

\paragraph{Contribution.} Our contribution are succinctly outlined as: 
\begin{enumerate}
\item We address the underlying fallacies of
Kernelized Convex Clustering (KCC), where data points are projected to a
Hilbert space $\mathcal{H}$, and subsequently, convex clustering is performed on
the projected data points. We propose an alternate algorithm that
leverages vanilla convex clustering itself to solve the problem
effectively. The convexity property of the optimization leads to a unique
minimizer, which we approximate after several iterations of our
Alternating Direction Method of Multipliers (ADMM) based algorithm
\citep{10.1561/2400000003}. As an interesting consequence, this method
naturally leads to an embedding in a finite lower-dimensional vector
space, whose convex clustering turns out to be equivalent to the kernel
convex clustering of the original data.

\item We study KCC from a theoretical aspect, establishing its convergence and
providing finite sample bounds on the iterates and the ground truths.
Further, the statistical properties of the finite-dimensional embedding
are vividly discussed. This analysis provides certain interesting insights
into its underlying structure and its relationship with the projected data
points. This aids in identifying patterns that can enhance both the
performance and interpretability of the model. We discuss some theoretical
details in Section~\ref{sec:theory} and provide extensive derivations in
Section 3 of the Appendix. Finally, we compare our method
with various state-of-the-art clustering algorithms and obtain impressive
performances on various benchmark datasets.
\end{enumerate}

\begin{figure}[!ht]
  \centering
     \includegraphics[width=0.40\linewidth]{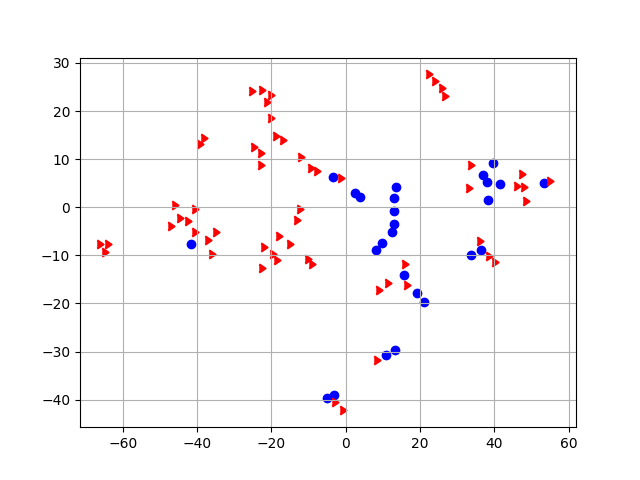} 
     \includegraphics[width=0.40\linewidth]{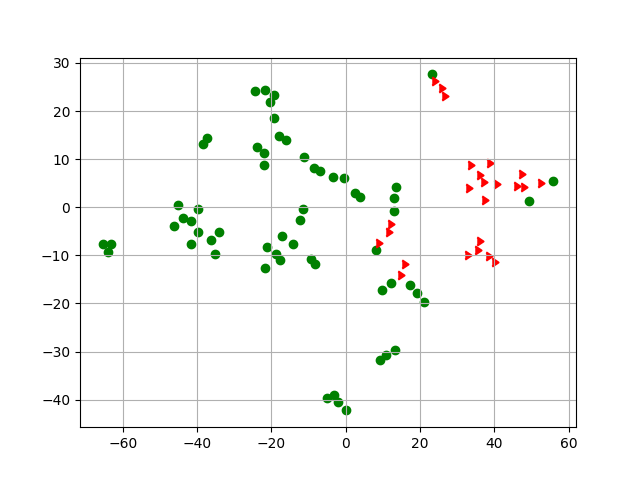} \\ 
     (a)  \hspace{0.33\linewidth} (b)\\
     \includegraphics[width=0.40\linewidth]{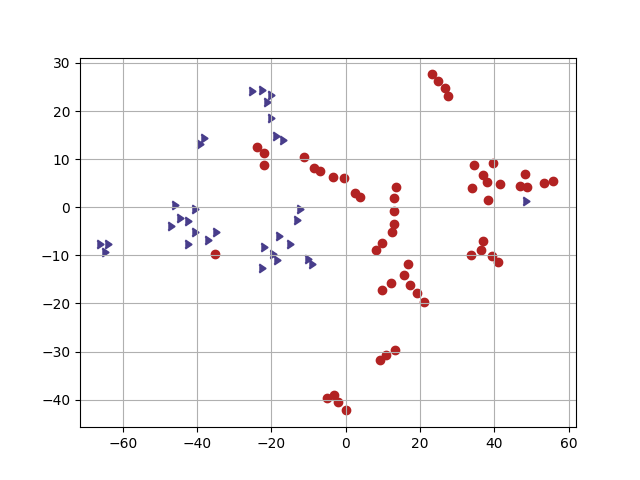}   
     \includegraphics[width=0.40\linewidth]{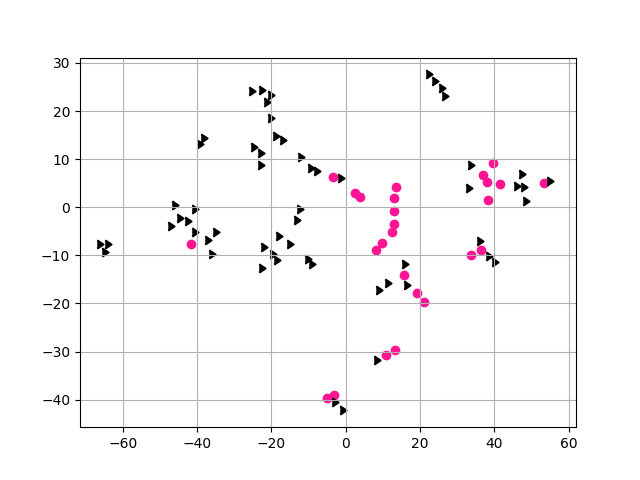} \\
     (c) \hspace{0.33\linewidth} (d) 
  \caption{t-SNE plots of GLI85 dataset for (a) ground truth labels,
    (b) $k$-means clustering, (c) convex clustering, and (d) KCC are
    presented. Applying kernels improves performance over the Euclidean
    similarity measure.}
  \label{fig:gli85}
\end{figure}

\section{Proposed Method}
The existing clustering algorithms, like $k$-means or convex clustering,
are inefficient for clustering data points that are not linearly separable
and contain non-convex patterns. The shortcomings can be alleviated by
pursuing kernel methods that project the data points into a
higher-dimensional Hilbert space, where data points are linearly
separable. This fact motivates us to design a kernelized clustering
algorithm to cluster intricately complex datasets.

\subsection{A Motivating Example}

We demonstrate our approach using the biological dataset, GLI85, which
comprises 85 samples and 22283 continuous features. Initially, the
dataset is pre-processed by standardizing the features. Refer to
Figure \ref{fig:gli85}a to observe the actual clusters present in GLI85.
Figures \ref{fig:gli85}b and \ref{fig:gli85}c aptly demonstrate that the
inefficiencies of $k$-means and convex clustering are due to their reliance on Euclidean-based similarity measures.
Furthermore, we respectively obtain 0.051 and 0.206 as the NMI values,
signifying the distortion of the cluster structure. In contrast,
kernelized convex clustering captures the accurate cluster structures as
evident in Figure~\ref{fig:gli85}d. In this context, we employed a
Gaussian kernel in our implementation as $k(\bm{x}, \bm{y}) =
e^{-\|\bm{x}-\bm{y}\|^2/2\sigma^2}$ where $\sigma$ was chosen to be
0.001. The NMI \citep{JMLR:v11:vinh10a} score was found to be 1 in this case, highlighting the
utility of the kernels in the paradigm of convex clustering.

\subsection{Problem Formulation}

Let $\{\bx_1,\bx_2, \dots, \bx_n\} \subseteq \mathbb{R}^d$ be $n$ data points to be
clustered. Let $\phi : \mathbb{R}^d \to \mathcal{H}$ be a feature map that maps every
data point $\bx_i$ to $\phi(\bx_i)$ in the Reproducing Kernel Hilbert Space,
$\mathcal{H}$. Let $\bu_i \in \mathcal{H}$ be the centroid corresponding to $\phi(\bx_i)$.
We propose to solve the following optimisation problem
\begin{equation}
  \min_{\bu_1,\dots,\bu_n}\ \frac{1}{2}\sum_{i=1}^{n}\|\phi(\bx_i)-\bu_i\|^2
  + \gamma \sum_{i < j} w_{ij}\|\bu_i - \bu_j\|
  \label{eq:kcc}
\end{equation}

Equation~\eqref{eq:kcc} has two separate summand terms. The first
summation is a measure of the fit of the model: the smaller this term is,
the closer the $\phi(\bx_i)$'s are to their corresponding centroids,
$\bu_i$'s, indicating a good fit of the model. The second term is a
penalisation term, to keep the number of distinct centroids in check. The
smaller this penalty term is, the fewer the number of distinct cluster
centroids. Here $\gamma$ is the tuning parameter for the fusion penalty
term $\sum w_{ij}\|\bu_i - \bu_j\|$, while $w_{ij}$'s are non-negative
weights for every pair of data points $i$ and $j$. $\gamma$ serves as a
tradeoff between the model fit and the model complexity. The larger
$\gamma$ is, the more probable it is that the cluster centroids fuse to
make the fusion penalty small, and thus minimise the entire objective. It
is a good choice to select the weights in a way that depends on the
proximity of $\bx_i$ and $\bx_j$. 

Associated with the map $\phi$ is an inner product $\langle\cdot,\cdot\rangle$
of the Hilbert space $\mathcal{H}$, which satisfies all three properties of an
inner product: symmetry, linearity, and positive-definiteness.
Accordingly, we also have the kernel function, $k : \mathbb{R}^d \times \mathbb{R}^d \to
\mathbb{R}^{+}$ such that $k(x_i, x_j) = \langle\phi(\bx_i),\phi(\bx_j)\rangle$, and
the kernel matrix $\bK$, whose $(i,j)^{\text{th}}$ entry is $k(\bx_i, \bx_j)$.
Define $\bm{\phi} = [\phi(\bx_1), \phi(\bx_2), \dots, \phi(\bx_n)]^\top$.
Note that $\bK = \bm{\phi}\bm{\phi}^\top$.

\subsection{Towards Optimization}
Fix $\bu_1, \dots, \bu_n \in \mathcal{H}$. Decompose each $\bu_i$ into the linear
space, $V = \mathrm{span}\{\phi(\bx_1), \phi(\bx_2), \dots, \phi(\bx_n)\}
\subseteq \mathcal{H}$ and its complement, $V^\perp$. Thus, for all
$i = 1, \dots, n$, $\exists\, \bm{\alpha}_i \in \mathbb{R}^n$ and $\bv_i \in {\bV}^\perp$,
such that
\[
  \bu_i = \bm{\phi}^\top \bm{\alpha}_i + \bv_i
\]
Now, observe that
\begin{align*}
  \|\phi(\bx_i) - \bu_i\|^2
  = \|\phi(\bx_i) - \bm{\phi}^\top\bm{\alpha_i} - \bv_i\|^2 
  = \|\phi(\bx_i) - \bm{\phi}^\top\bm{\alpha_i}\|^2 + \|\bv_i\|^2 
  \geq \|\phi(\bx_i) - \bm{\phi}^\top\bm{\alpha_i}\|^2
\end{align*}
In the second equality, there is no term of the inner product because
$\phi(\bx_i) - \bm{\phi}^\top\bm{\alpha}_i$ and $\bv_i$ are orthogonal. The inequality becomes an equality if and only if $\bv_i = 0$. Similarly, for
each of the different terms in the second summation,
\begin{align*}
  \|\bu_i - \bu_j\|^2
  = \|\bm{\phi}^\top(\bm{\alpha}_i - \bm{\alpha}_j) + \bv_i - \bv_j\|^2 
  = \|\bm{\phi}^\top(\bm{\alpha}_i - \bm{\alpha}_j)\|^2 + \|\bv_i - \bv_j\|^2 
  \geq \|\bm{\phi}^\top(\bm{\alpha}_i - \bm{\alpha}_j)\|^2
\end{align*}
Thus,
\begin{align*}
    \|\bu_i - \bu_j\|\geq \|\bm{\phi}^\top(\bm{\alpha}_i - \bm{\alpha}_j)\| 
  \implies \sum_{i<j}\|\bu_i - \bu_j\| \geq 
  \sum_{i<j}\|\bm{\phi}^\top(\bm{\alpha}_i - \bm{\alpha}_j)\|
\end{align*}
Combining all these, we get the value of \ref{eq:kcc} at
$\bu_1, \dots, \bu_n$ is greater than or equal to at $\bm{\phi}^\top
\bm{\alpha}_1, \dots, \bm{\phi}^\top\bm{\alpha}_n$ and equality holds if and only
if $\bv_1 = \dots = \bv_n = 0$. Hence, if $\bu^*_1, \dots, \bu^*_n$'s are the
minimisers in \ref{eq:kcc}, their projections $\bv^*_i \in V^\top$ must
all equal the zero vector. Thus $\bu^*_i = \bm{\phi}^\top\bm{\alpha}^*_i$ for
some $\bm{\alpha}^*_i \in \mathbb{R}^n$. This observation turns out to be helpful, as
we can just substitute $\bm{\phi}^\top\bm{\alpha}_i$ for every $\bu_i$ in
Equation \ref{eq:kcc} and try to minimise it with respect to
$\bm{\alpha}_1, \dots, \bm{\alpha}_n$.

\begin{remark}\normalfont
Although the above theorem is similar to the familiar representer theorem
of Hilbert Spaces \citep{Shalev-Shwartz_Ben-David_2014}, it is quite
different from the problem statement in \ref{eq:kcc}, and how it is
derived. Firstly, the representer theorem seeks to find a minimiser of a single
variable in $\mathcal{H}$; in this case, we have $n$ such unknowns, i.e.\ $\bu_i$'s.
More importantly, the representer theorem involves a penalty term which is
strictly increasing in the norm of the unknown variable, contrary to our
case, where the penalty term involves the norm of the pairwise differences.
$\|\bu_i - \bu_j\|$. Clearly, the penalty is not increasing in the norm of
the unknown variables, $\|\bu_i\|$'s.
\end{remark}

\subsection{A Perspective from Convex Clustering}
Substituting $\bu_i=\bm{\phi}^\top\bm{\alpha}_i$, and recalling that
$\bK = \bm{\phi}\bm{\phi}^\top, \phi(\bx_i)=\bm{\phi}^\top\be_i$, we see
\begin{align*}
  \|\phi(\bx_i)-\bm{\phi}^\top\bm{\alpha}_i\|^2
   = (\bm{\alpha}_i-\be_i)^\top \bK (\bm{\alpha}_i-\be_i)
  \|\bu_i - \bu_j\|^2
  = (\bm{\alpha}_i-\bm{\alpha}_j)^\top \bK (\bm{\alpha}_i-\bm{\alpha}_j)
\end{align*}
The reformulated problem, being convex in $\bm{\alpha}_i$'s, can be solved by computational techniques like
ADMM or AMA \citep{10.1561/2400000003,Chi_2015}. However, to avoid computational challenges, we resort
to a different approach. If we decompose $\bK = \bZ^\top \bZ$ using Cholesky
decomposition, and make the following transformations:
\begin{equation}
  \bz_i = \bZ \be_i, \quad \ba_i = \bZ\bm{\alpha}_i
  \label{eq:transform}
\end{equation}
We get a transformed objective function:
\begin{equation}
  \min_{\ba_1,\dots,\ba_n}\ \frac{1}{2}\sum_{i=1}^{n}\|\bz_i - \ba_i\|^2
  + \gamma \sum_{i < j} w_{ij}\|\ba_i - \ba_j\|
  \label{eq:transformed}
\end{equation}
which is the objective for the convex clustering of the $n$ points,
$\bz_1, \dots, \bz_n$~\eqref{eq:convex_clust}. Cholesky decomposition of
the kernel matrix $\bK = \bZ^\top \bZ$ aids us in reducing KCC to the
well-known convex clustering problem. So, solving the kernel convex
clustering problem in Equation \ref{eq:kcc} simultaneously leads to an
embedding of the $n$ points in $\mathbb{R}^n$, whose convex clustering is
equivalent to KCC in \ref{eq:kcc}.

\subsection{Algorithms for Kernelized Convex Clustering}
\begin{algorithm}[h]
\caption{\textsc{Kernelized Convex Clustering} (KCC)}
\label{alg:kcc}
\begin{algorithmic}
\REQUIRE $\bx_1, \dots, \bx_n \in \mathbb{R}^d$,\ $k(\cdot,\cdot)$,\ $w_{ij}$,\
         $\rho, \gamma > 0$
\STATE Initialise $\bm{\alpha}_i = \be_i$ for all $i = 1, \ldots, n$
\STATE Initialise $\bm{\beta}_{ij} = \be_i - \be_j$ and $\bm{\lambda}_{ij}$ for all
       $i < j$ such that $w_{ij} > 0$
\WHILE{does not converge}
  \STATE $\bm{\alpha}^{(m)}_i =
    \frac{\be_i + \bm{\tau}_{f}^{(m-1)}
    - \bm{\tau}_{b}^{(m-1)}}{1 + n\rho}
    + \frac{\rho\sum_{i=1}^n \be_i}{1 + n\rho}$
  \STATE $\bm{\lambda}^{(m)}_{ij} = \bm{\lambda}^{(m-1)}_{ij}
    + \rho \bK(\beta^{(m-1)}_{ij}
    - \bm{\alpha}^{(m)}_i + \bm{\alpha}^{(m)}_j)$
  \STATE $\bm{\beta}^{(m)}_{ij} =
    \left(1 - \frac{\sigma_{ij}}{\sqrt{\bt^{(m)\top}_{ij}
    \bK \bt^{(m)}_{ij}}}\right)_{+}(\bm{\alpha}^{(m)}_i - \bm{\alpha}^{(m)}_j
    - \bK^{-1}\bm{\lambda}^{(m)}_{ij}/\rho)$
   \STATE $\bm{\tau}_{f}^{(m)}=\sum_{j=1}^{n}(\bK^{-1}\bm{\lambda}^{(m)}_{ij}
    + \rho\bm{\beta}^{(m)}_{ij})$ 
    \STATE 
    $ \bm{\tau}_{b}^{(m)}=\sum_{j=1}^{n}(\bK^{-1}\bm{\lambda}^{(m)}_{ji}
    + \rho\bm{\beta}^{(m)}_{ji})$
\ENDWHILE
\end{algorithmic}
\end{algorithm}

\cite{Chi_2015} proposed two splitting
methods for convex clustering, one using the Alternating Direction Method
of Multipliers (ADMM) \citep{10.1561/2400000003} and the other one using the Alternating Minimization Algorithm (AMA). Since ADMM converges under broader conditions than AMA (Section 4 of \cite{Chi_2015}), we have used the former one to get updates of the
$\ba_i$'s; then we revert the transformations in Equation~\eqref{eq:transform}
to get the solution of the $\bu_i$'s. In ADMM, we introduce auxiliary
variables, $\bv_{ij} = \ba_i - \ba_j$, which act as constraints, when we
rewrite \ref{eq:transformed} by replacing $\ba_i - \ba_j$ with $\bv_{ij}$,
and optimise it with respect to the $\ba_i$'s and $\bv_{ij}$'s. Additionally,
we also introduce Lagrange multipliers $\bm{\eta}_{ij}$ corresponding to
$\bv_{ij}$, and a hyperparameter $\rho > 0$, which controls the effect of
the quadratic penalty term $\sum_{i<j}\|\bv_{ij} - \ba_i + \ba_j\|^2$ in
the ADMM objective.
{\small
\begin{align*}
  \ba_i &= \frac{\bz_i + \sum_{j=1}^{n}(\bm{\eta}_{ij} + \rho \bv_{ij})
               - \sum_{j=1}^{n}(\bm{\eta}_{ji} + \rho \bv_{ji})}{1 + n\rho}
         + \frac{\rho\sum \bz_i}{1+n\rho}\\[4pt]
  \bm{\eta}_{ij} &= \bm{\eta}_{ij} + \rho(\bv_{ij} - \ba_i + \ba_j)\\[4pt]
  \bv_{ij} &= \left(1 - \frac{\sigma_{ij}}{\norm{\ba_i - \ba_j - \eta_{ij}/\rho}}\right)_{\!\!+}
            (\ba_i - \ba_j - \bm{\eta}_{ij}/\rho)
\end{align*}}
In the last equation, $\sigma_{ij} = \frac{\gamma w_{ij}}{\rho}$. Now
note that $\bK = \bZ^\top \bZ$ and $\bK^{-1} = \bZ^{-1}{\bZ^{\top}}^{-1}$. We could invert
$\bZ$, because almost surely the data to be clustered will come from a
continuous distribution, making $\bK$ non-singular. Letting $\bm{\lambda}_{ij} =
\bZ^\top \bm{\eta}_{ij}$, $\bv_{ij} = \bZ\bm{\beta}_{ij}$ and recalling that
$\ba_i = \bZ\bm{\alpha}_i$, we can rewrite the updates for $\bm{\alpha}_i$, $\bm{\beta}_{ij}$,
$\bm{\lambda}_{ij}$. The required updates can be found in
Algorithm \ref{alg:kcc}.

\begin{remark}\normalfont
We see that KCC of a dataset with kernel matrix $\bK$, is equivalent to
convex clustering of the embedded matrix $\bZ$, which satisfies
$\bZ^\top \bZ = \bK$. We can choose $\bZ$ in any way possible as long as it
satisfies the above condition. Using Cholesky decomposition makes $\bZ$
upper triangular, giving the embedding a redundant structure. However, not
all embeddings may have a redundant structure. To see this, suppose $\bZ$
is a suitable embedding. We select an orthogonal matrix $\bQ$, so that $\bQ\bZ$
is neither upper nor lower triangular. Since $(\bQ\bZ)^\top \bQ\bZ = \bZ^\top \bZ =
\bK$, $\bQ\bZ$ is also an embedding. This further demonstrates that the
embedding is not unique. The number of embeddings is infinite, because of
the possible infinite choices of the orthogonal matrix $\bQ$.
\end{remark}
After getting the embedding $\bZ$, one can use any convex clustering method
to get the $\ba_i$'s. A similar AMA algorithm can be derived in a fashion
similar to Algorithm \ref{alg:kcc}, using the steps mentioned in
\cite{Chi_2015}. Other notable methods to
convex cluster $\bZ$ include Cluster-path as mentioned in
\cite{inproceedings}. Since ADMM-based
convex clustering converges, it also guarantees the convergence of KCC.

\subsection{Getting the Optimal Number of Clusters}

The final step involves determining the optimal number of clusters and the
corresponding cluster assignments of the data points. This is carried out,
first by applying agglomerative clustering on the centroids, followed by
constructing a dendrogram. Now, for a given number of clusters $k$, the
dendrogram is cut at a suitable height to obtain $k$ clusters and get the
respective labels. For this $k$, we compute the fit of the data using the
standard $k$-means sum of squares formula: 
$SSE_k = \sum_{t=1}^{k}\sum_{i \in C_k}
  \left\|\hat{\bu}_i - \frac{\sum_{j \in C_t}\hat{\bu}_j}{|C_t|}\right\|^2$.
After computing $SSE_k$ for every $k$, we construct the elbow
plot of $\mathrm{SSE}_k$ vs.\ $k$. We identify the elbow point as the
point after which the change in $SSE_k$ becomes small with
respect to previous changes, thereafter. In other words, the graph
continues to be approximately linear afterwards with the same slope for a
long range of values. We also expect this slope not to be quite big. The
value of $k$, corresponding to this elbow point, denotes the optimal
number of clusters for the dataset.
\subsection{Complexity Analysis}

In KCC, the storage complexity is $O(n^2)$ for first storing the kernel
matrix $\bK$, and an additional $O(n^2)$ for storing the vectors $\bm{\alpha}_i$.
So the total storage complexity in this case is $O(n^2)$. In comparison,
kernel power $k$ means (KPKM) \citep{9928792} has storage complexity
$O(n^2)$, and that of biconvex clustering (BCC) \citep{chakraborty2021biconvexclustering}
is $O(np)$. In case of high-dimensional data, with $p \gg n$, KCC turns
out to be better than biconvex clustering in terms of memory requirements.
In terms of computational complexity, KCC takes $O(n^3)$ number of
operations, KPKM takes $O(n^2k + npk)$, while BCC takes $O(n^2p)$. When
comparing with KPKM, there is a tradeoff between cluster number and
dimensionality, since we need to give the number of clusters $k$ as input.
Also, in both cases, the dimensionality plays a crucial role in the
complexity. For high-dimensional datasets again with $p \gg n$, KCC
overpowers BCC. For KPKM, although the complexity is lower than KCC, KCC
predicts the actual number of clusters using the elbow plot. Thus in
arbitrarily shaped datasets, KPKM may not give a proper clustering with a
given $k$, but KCC automatically predicts the actual number of clusters.

\section{Theoretical Properties}
\label{sec:theory}

In this section, we will offer insights on the finite sample properties of
the estimates and the consistency of the algorithm. Let $\bu_i$ be the
ground truth corresponding to the centroid of the $i^{\text{th}}$ data
point, and $\hat{\bu}_i$ be the centroid estimates by minimising
Equation \ref{eq:kcc}. The square loss term, $\sum_{i=1}^{n}
\|\phi(\bx_i) - \bu_i\|^2$, in \ref{eq:kcc}, measures the fit of the
data with the ground truths. The more close $\hat{\bu}_i$ and $\bu_i$ are,
the more close the two quantities $\sum_{i=1}^{n}\|\phi(\bx_i) -
\bu_i\|^2$ and $\sum_{i=1}^{n}\|\phi(\bx_i) - \hat{\bu}_i\|^2$ become, and
the better is the fit of the data. So it makes sense to bound the norm
$\sum_{i=1}^{n}\|\hat{\bu}_i - \bu_i\|^2$. We assume the following data
generating process:
\[\phi(\bx_i) = \bu_i + \bm{\epsilon}_i\]
for all data points $i = 1, \dots, n$. Here $\phi(\bx_i)$, $\bu_i$ and
$\bm{\epsilon}_i$ are all assumed to lie in the Hilbert Space $\mathcal{H}$.
$\bm{\epsilon}_i$ are mean zero sub-Gaussian random variables in $\mathcal{H}$ with
respect to an operator $\Gamma$ \citep{chen2020hansonwrightinequalityhilbertspaces}. Further, assume that $\bm{\epsilon}_i$'s are independent, thus implying that
$\mathbb{E}[\langle\bm{\epsilon}_i, \bm{\epsilon}_j\rangle] = 0$ for distinct $i$
and $j$. Also, suppose $\mathbb{E}[\langle\bm{\epsilon}_i,\bm{\epsilon}_i\rangle] =
\sigma^2$. The sub-Gaussian model assumptions are fairly standard and are
quite analogous to the ones mentioned in
\cite{tan2015statisticalpropertiesconvexclustering}
 and \cite{Wang_2018}. For practical purposes, we
can assume that the error terms $\bm{\epsilon}_i$'s are bounded in $\mathcal{H}$, that
is $\|\bm{\epsilon_i}\| \leq M$ almost surely, for some $M > 0$ and $i =
1, \ldots, n$. The bounded support assumption is standard in the theoretical analysis for centroid-based clustering \citep{pollard1981strong,paul2021uniform,paul2021neurips}.  This is surely the case if we take our kernel function
$k(\bx,\by)$ to be the radial basis function, $k(\bx,\by) =
\exp(-\|\bx-\by\|^2/2\sigma^2)$. Observe that,
\begin{align*}
  \|\phi(\bx_i) - \bu_i\|^2 - \|\phi(\bx_i) - \hat{\bu}_i\|^2
  =  -\|\bu_i - \hat{\bu}_i\|^2 - 2\langle\bm{\epsilon}_i, \bu_i - \hat{\bu}_i\rangle
\end{align*}
Further, \eqref{eq:kcc} is minimized by $\hat{\bu}_i$, and so the first quantity in the above string of equalities can be lower bounded by a difference of the penalty terms at $\hat{\bu}_i$ and $\bu_i$. Thus, it
remains to bound only $\langle\bm{\epsilon}_i, \bu_i - \hat{\bu}_i\rangle$. We
accomplish this by decomposing $\mathcal{H}$ into two mutually orthogonal spaces
in $\mathcal{H}$ and analyzing the projections of the estimates in the orthogonal
subspaces, separately. The details of our calculations are mentioned
explicitly in the Appendix. We summarise the statistical
analysis in the following theorem.

\begin{theorem}
\label{thm:main}
Let $\phi(\bx_i) = \bu_i + \bm{\epsilon}_i$ for al $i = 1, \dots, n$, where
$\bm{\epsilon}_i$ are i.i.d.\ mean zero sub-Gaussian random variables in the
RKHS $\mathcal{H}$, with respect to the operator $\Gamma$. Let $\hat{\bu}_i$ be
the solutions of \ref{eq:kcc}. If $\gamma \geq \frac{2z_0}{w_{\min}}$,
then
\[
  \frac{1}{2n}\sum_{i=1}^{n}\|\hat{\bu}_i - \bu_i\|^2
  \leq \frac{3\gamma}{2n}\sum_{i<j}w_{ij}\|\bu_i - \bu_j\|
  + \sigma^2\!\left[\frac{1}{n} + \sqrt{\frac{\log(n)}{n^2}}\right]
\]
with probability at least $1 - \dfrac{2}{\binom{n}{2}} -
2\exp\!\left[-C\min\!\left(\dfrac{\sigma^4\log(n)}{L^4\|\Gamma\|^2_{\mathrm{HS}}},\,
\dfrac{\sigma^2\sqrt{\log(n)}}{L^2\|\Gamma\|^2_{\mathrm{op}}}\right)\right]$
for some constant $C$.
\end{theorem}

The above theorem is proved by using Hanson-Wright's Inequality \citep{chen2020hansonwrightinequalityhilbertspaces}
 on each of the
mutually orthogonal projections. The term $\bz_0$ is explicitly described
in the Appendix; it is dependent on the data through
$\sigma$ and $\Sigma$, and is $O(\log n)$. Thus, the sub-Gaussianity assumption greatly reduces the data dependency of $\gamma$ to just
$\sigma$ and $\Sigma$, to get the above finite sample bounds.

\begin{remark}\normalfont
In Theorem~\ref{thm:main}, the fusion parameter is dependent on the value
of $\bz_0$, to attain this upper bound. Now, if $\|\bu_i - \bu_j\|$'s are
uniformly bounded for all pairs $i \neq j$, and $\frac{\gamma}{n}
\sum_{i<j}w_{ij} = o_p(1)$ as $n \to \infty$, then the right hand side
of the inequality goes to zero, with the probability of the event going
to 1. Thus, the average fit of the centroids goes to zero in such
circumstances. However, $\gamma \geq 2z_0/w_{\min}$ as stated in
Theorem~\ref{thm:main}. Thus, a necessary condition for $\frac{\gamma}{n}
\sum_{i<j}w_{ij} = o_p(1)$ to hold is to have $z_0\sum_{i<j}w_{ij} /
nw_{\min} \to 0$ as $n \to \infty$. From the definition of $z_0$, we see that it is at most of order
$O(\log n)$. Notice that there are exactly $\binom{n}{2}$ possible
$w_{ij}$'s. Suppose for some $0 < \alpha < \frac{1}{2}$, at most $n^\alpha$
of these $w_{ij}$'s are positive and the remaining ones are zero.
Rigorously stating, the number of elements in the set $\{w_{ij} : w_{ij}
> 0,\; i < j\}$ is less than or equal to $n^\alpha$, $0 < \alpha <
\frac{1}{2}$. Recall that $w_{\min} = \min\{w_{ij} : w_{ij} > 0,\;
i < j\}$. Further, suppose $\frac{1}{\sqrt{n}} \leq w_{ij} \leq 1$ for
all the weights lying in the aforementioned set. Then, $z_0\sum_{i<j}w_{ij} / nw_{\min}
  \leq z_0 n^\alpha / \sqrt{n}
  \leq 2\,O(\log n)\,/\,n^{\frac{1}{2}-\alpha} \to 0$.
Thus Algorithm~\ref{alg:kcc} is consistent if the above-mentioned
conditions hold simultaneously.
\end{remark}

Remark 3 gives us certain conditions on the weights $w_{ij}$, so that
our estimated centroids converge to the ground truth in the case of a
large number of samples. Specifically, we need to ensure that out of all
the $\binom{n}{2}$ $w_{ij}$'s possible, at most $\sqrt{n}$ of them must
be positive and the remaining should be set to zero. Further, the weights
should be within the range of $\frac{1}{\sqrt{n}}$ and 1. This is
practically possible, since $w_{ij}$'s are hyperparameters and can be
tuned accordingly by the practitioner. 

\begin{remark}\normalfont
For the bounds stated in Theorem~\ref{thm:main} to hold, the tuning
parameter $\gamma$ must be greater than $2z_0/w_{\min}$, where $z_0$
itself is a quantity dependent on $\mathcal{H}$. So the choice of the kernel
space indeed does affect the quality of clustering. Compared to Lemma 7
of \cite{tan2015statisticalpropertiesconvexclustering}
, where
$\gamma = \Omega\!\left(\sqrt{\log pn^2/np}\right)$, in our case,
$\gamma = \Omega(\sqrt{\log n})$ for similar kinds of bound to hold.
\end{remark}
\begin{table*}[t]
  \centering
  \caption{NMI scores of KCC and other clustering methods applied on
    different datasets.}
  \label{tab:benchmarks}
  \scriptsize
  \resizebox{0.90\columnwidth}{!}{
  \begin{tabular}{lcccccccccr}
    \toprule
    \textbf{Datasets}
      & \textbf{KCC (Ours)}
      & \textbf{Convex}
      & \textbf{$k$-Means}
      & \textbf{Spectral}
      & \textbf{KPKM}
      & \textbf{BCC}
      & \textbf{TGCC}
      & \textbf{C-PAINT}
      & \textbf{SC-NR}
      & \textbf{\#clusters} \\
    \midrule
    Lymphoma    & \textbf{0.778} & 0.718 & 0.654 & 0.179 & 0.633 & 0.450 & 0.727 & 0.689 & 0.429 &  7 \\
    Orlraws10P  & \textbf{0.851} & 0.821 & 0.798 & 0.209 & 0.810 & 0.720 & 0.720 & 0.759 & 0.645 & 11 \\
    Yale        & \textbf{0.657} & 0.293 & 0.480 & 0.601 & 0.568 & 0.288 & 0.549 & 0.617 & 0.477 & 14 \\
    Lung        & \textbf{0.804} & 0.729 & 0.594 & 0.018 & 0.433 & 0.328 & 0.769 & 0.682 & 0.321 &  4 \\
    Zoo         & \textbf{0.736} & 0.324 & 0.690 & 0.637 & 0.459 & 0.695 & 0.629 & 0.699 & 0.645 &  4 \\
    Housevotes  & \textbf{0.573} & 0.004 & 0.536 & 0.542 & 0.518 & 0.489 & 0.525 & 0.429 & 0.538 &  2 \\
    Glass       & \textbf{0.439} & 0.255 & 0.357 & 0.367 & 0.347 & 0.308 & 0.398 & 0.402 & 0.384 &  9 \\
    New Thyroid & \textbf{0.706} & 0.491 & 0.553 & 0.491 & 0.376 & 0.407 & 0.568 & 0.467 & 0.303 &  5 \\
    Glioma      & \textbf{0.529} & 0.506 & 0.490 & 0.031 & 0.411 & 0.453 & 0.451 & 0.279 & 0.091 &  3 \\
    MNIST       & \textbf{0.614} & 0.062 & 0.553 & 0.047 & 0.486 & 0.421 & 0.329 & 0.496 & 0.425 & 10 \\
    \bottomrule
  \end{tabular}}
\end{table*}

\section{Experiments}
In this section, we compare KCC~\eqref{alg:kcc} with various baseline and state-of-the-art clustering methods. We use Normalized Mutual Information (NMI) \citep{JMLR:v11:vinh10a} to evaluate the quality of clustering. In the following subsections, we establish KCC's superiority by applying it to several synthetic and real-world datasets and comparing it with various state-of-the-art methods. The additional experiments on hyperparameter sensitivity and various kernels are presented respectively in Sections \ref{sec:app_hyp_sens} and \ref{sec:app_diff_ker} of the Appendix. 

\subsection{Results on Synthetic Dataset}\label{Results on Synthetic Dataset}
\begin{figure}[!h]
  \centering
    \includegraphics[width=0.40\linewidth]{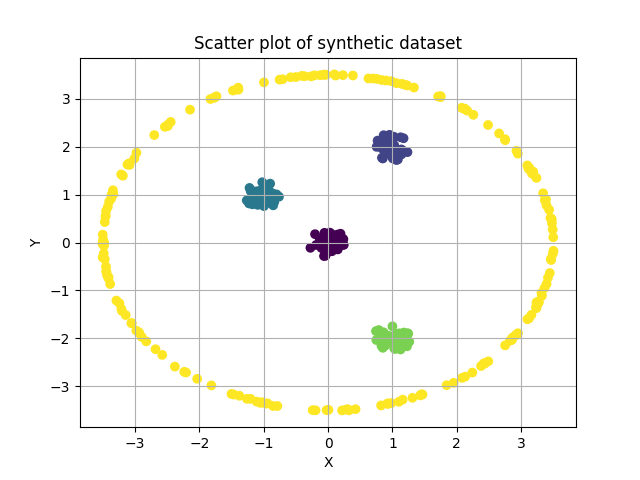}   
    \includegraphics[width=0.40\linewidth]{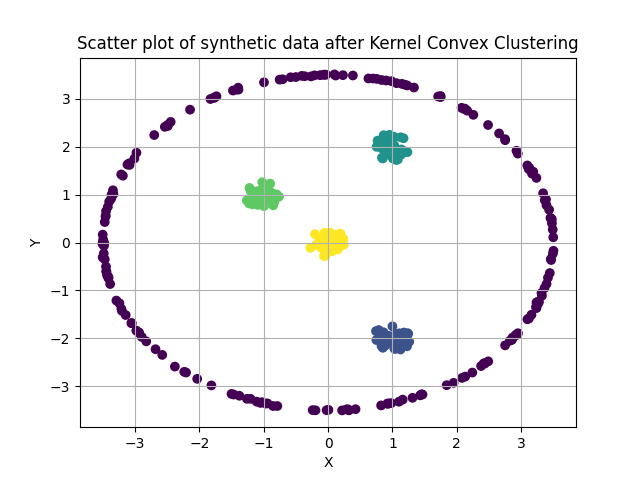} \\
    (a) \hspace{0.33\textwidth} (b) \\
    \includegraphics[width=0.40\linewidth]{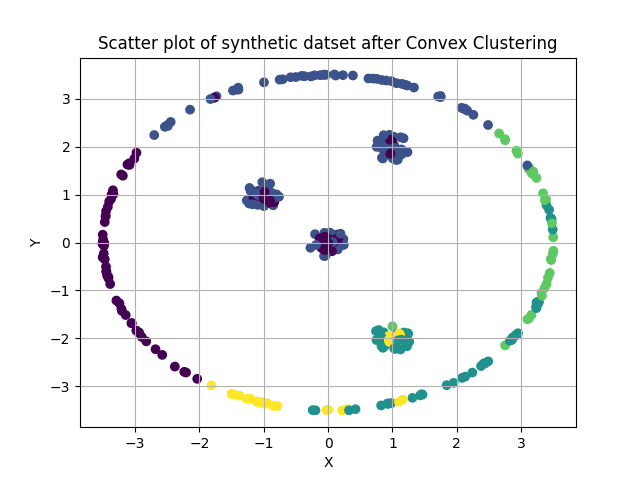}
    \includegraphics[width=0.40\linewidth]{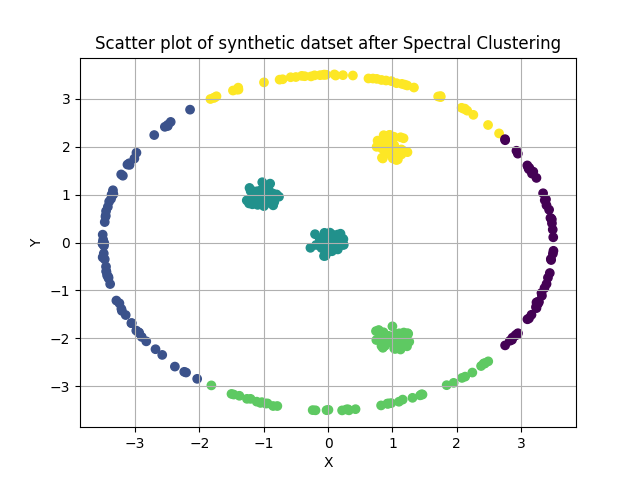} \\
    (c) \hspace{0.33\textwidth} (d) \\
  \caption{Scatter plots of the synthetic dataset for (a) ground truth
    labels, (b) KCC, (c) convex clustering, and (d) spectral clustering
    are illustrated.}
  \label{fig:synthetic}
\end{figure}

\begin{table}[!ht]
  \centering
  \scriptsize
  \caption{NMI values after applying different methods on the synthetic
    dataset.}
  \label{tab:synthetic}
  \small
  \begin{tabular}{lc}
    \toprule
    \textbf{Method} & \textbf{NMI} \\
    \midrule
    Kernel Convex Clustering & 0.999 \\
    Convex Clustering        & 0.259 \\
    Biconvex clustering      & 0.721 \\
    Kernel Power $k$ means   & 0.448 \\
    TGCC                     & 0.784 \\
    C-PAINT                  & 0.582 \\
    SC-NR                    & 0.671 \\
    Spectral Clustering      & 0.598 \\
    $k$-means                & 0.457 \\
    \bottomrule
  \end{tabular}
\end{table}
We generate a simulated dataset of 400 data points in $\mathbb{R}^2$, as shown
in Figure~\ref{fig:synthetic}. The four central blobs each consist of 50
points, while the outer circle comprises 200 points. For simulating each
of the blobs, first we generate $\theta_i \overset{\text{i.i.d.}}{\sim}
U(0, 2\pi)$, $R_i \overset{\text{i.i.d.}}{\sim} U(0, 0.45)$. Then we
set $x_i = R_i\cos(\theta_i)$ and $y_i = R_i\sin(\theta_i)$. We
accordingly shift the points to finally get 4 such blobs of size 50 each.
Next, we again generate $\theta_i \overset{\text{i.i.d.}}{\sim}
U(0, 2\pi)$, and set $x_i = 3\cos(\theta_i) + \epsilon_{i1}$ and
$y_i = 3\sin(\theta_i) + \epsilon_{i1}$, where $\epsilon_{i2},
\epsilon_{i2} \overset{\text{i.i.d.}}{\sim} \mathcal{N}(0, 0.01)$.



\begin{figure}[!ht]
    \centering
    \begin{subfigure}[t]{0.48\linewidth}
        \centering
        \includegraphics[width=\linewidth]{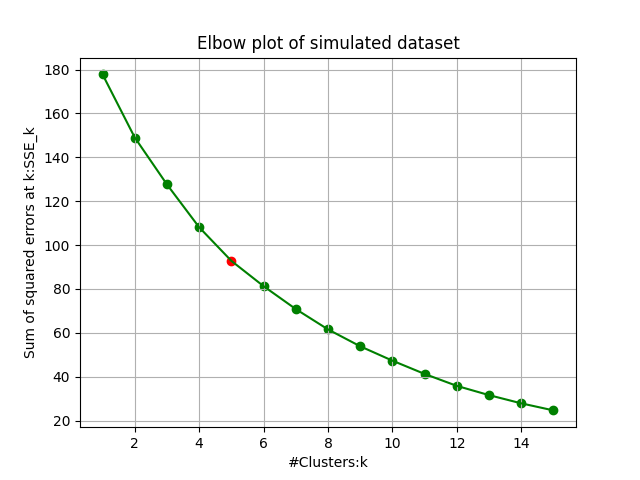}
        \caption{}
    \label{fig:elbow_syn}
    \end{subfigure}
    \hfill
    \begin{subfigure}[t]{0.48\linewidth}
        \centering
        \includegraphics[width=\linewidth]{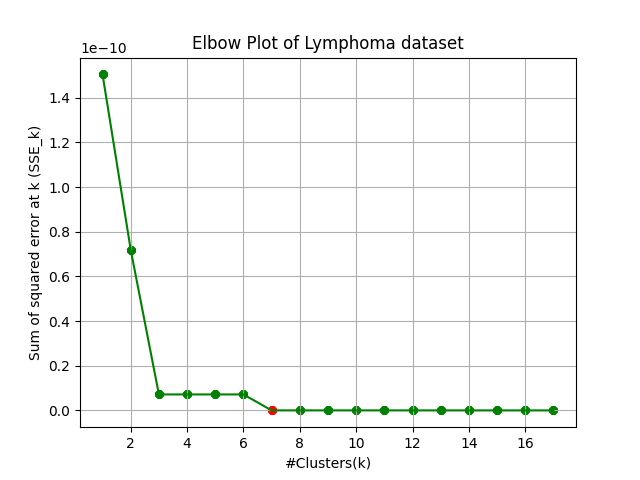}
        \caption{}
        \label{fig:elbow_lymphoma}
    \end{subfigure}
    \caption{In (a) Elbow plot of synthetic dataset with $\sigma_1 = 1$,
    $\sigma_2 = 100$, $\gamma = 1$, $\rho = 0.001$. This set of values
    gives the optimal clustering with NMI of 1, and in (b), the elbow plot for the Lymphoma dataset. The optimal number of clusters is estimated to be $7$. The dataset contains $9$ annotated clusters, although several contain very few samples and are therefore merged by KCC. }
    \label{fig:combined_elbow_plot}
\end{figure}

In this way, we get the outer circle. We use the Gaussian kernel as the
feature map to project the 400 points in an RKHS $\mathcal{H}$, which is a
popular choice for the feature map. The kernel function associated with
it is $k(\bm{x}, \bm{y})=e^{-\|\bm{x}-\bm{y}\|/2\sigma_1^2}$. The
weights were chosen as follows: for every pair $i \neq j$,
$w_{ij} = e^{-\|x_i - x_j\|^2/2\sigma_2^2}\,
\mathbb{I}[x_j \text{ is one of the 6 nearest neighbours of } x_i]$.
To make the weights symmetric, we finally chose $w^*_{ij} = (w_{ij} +
w_{ji})/2$. $\rho$ and $\gamma$ were also chosen after proper tuning.

\begin{figure}[!ht]
  \centering
  \includegraphics[width=0.60\linewidth]{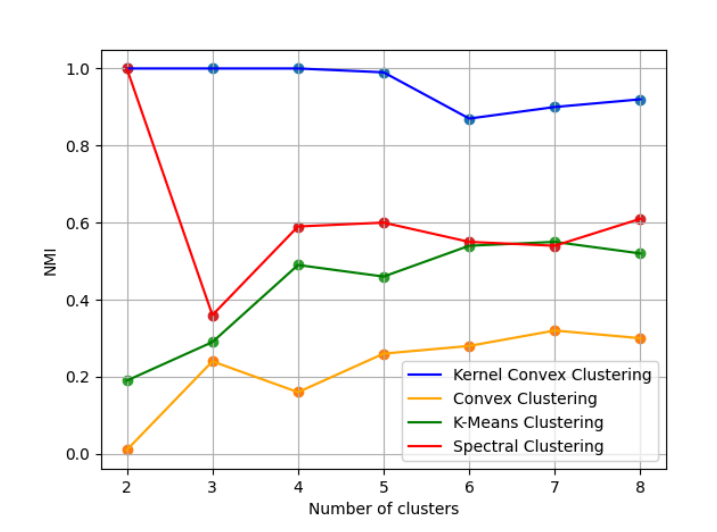}
  \caption{The impact on NMI with varying numbers of clusters is
    presented. Our method KCC performs consistently compared to other
    methods.}
  \label{fig:nmi_vs_k}
\end{figure}

We further demonstrate the result of other competing methods, like convex
clustering, spectral clustering, kernel $k$-means, kernel power $k$ means
\citep{9928792}, biconvex clustering \citep{chakraborty2021biconvexclustering},
C-PAINT \cite{pmlr-v161-zhang21d}, SC-NR \citep{unknown} and TGCC
\citep{zhang2025treeguidedl1convexclustering}. The scatter plots
corresponding to ground truths, KCC, convex clustering and spectral
clustering are elucidated in Figure~\ref{fig:synthetic}. The
corresponding NMI values are also reported in
Table~\ref{tab:synthetic}, and the corresponding elbow plot is
demonstrated in Figure~\ref{fig:elbow_syn}.

\begin{figure*}[!ht]
  \centering
    \includegraphics[width=0.23\linewidth]{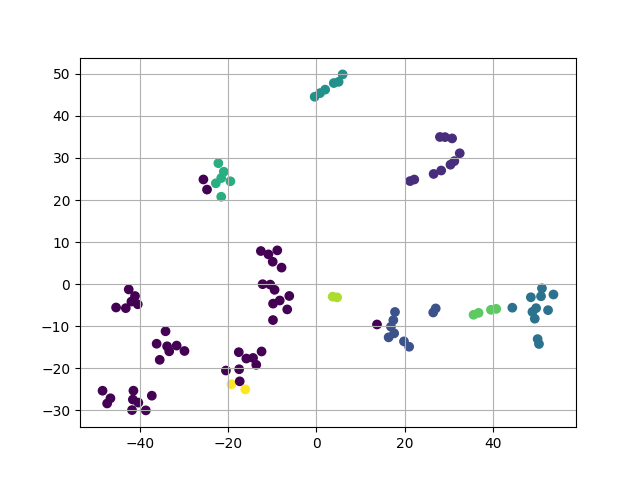} 
    \includegraphics[width=0.23\linewidth]{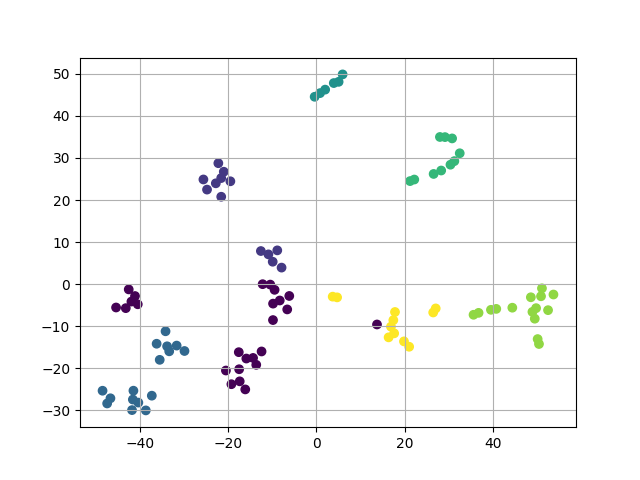}
    \includegraphics[width=0.23\linewidth]{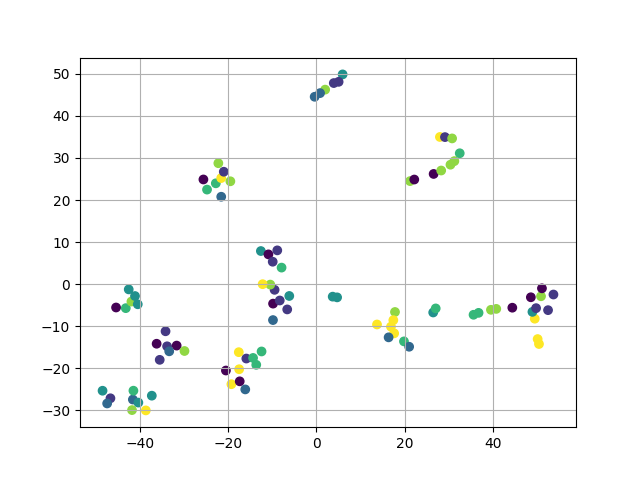}  
    \includegraphics[width=0.23\linewidth]{tsne_lymphoma_spectral.png} \\
    (a) \hspace{0.20\textwidth} (b) \hspace{0.20\textwidth} (c) \hspace{0.20\textwidth} (d) \\
  \caption{t-SNE plots of Lymphoma dataset for (a) ground truth labels,
    (b) KCC, (c) spectral clustering, and (d) $k$-means clustering are
    presented.}
  \label{fig:lymphoma_tsne}
\end{figure*}
\subsection{Effect of Increasing Number of Clusters}

We now check the efficacy of our method on an increasing number of
clusters. The number of clusters varies from two to eight. We use the
same kind of synthetic dataset as used in the previous experiment. For
$k = 2$ clusters, we have the outer circle of 200 points and the central
blob of 50 points. As $k$ increases, blobs of 50 points are added one by
one inside the interior of the outer circle. The blobs and the outer
circle were generated in the manner described in subsection 4.1. On each of the datasets, we apply KCC and different clustering methods,
and compare the results using the NMI score. We graphically summarise the
effect of increasing the number of clusters on the NMI score in
Figure~\ref{fig:nmi_vs_k}. KCC turns out to be the best choice for
clustering as $k$ increases. The cluster predictions also turn out to be
mostly true for KCC in comparison to other methods.
\subsection{Case Study on Lymphoma Dataset}\label{Case Study on Lymphoma Dataset}

We assess our algorithm's performance by applying KCC on the Lymphoma
microarray dataset \cite{li2018feature}. It comprises 96 instances and
4026 features, all of which are discrete. In total, there were 9 classes,
two or three of which had very few instances belonging to them. Since the
variables were all discrete, we did not standardise the dataset.


We used the Gaussian kernel as the feature map. The weights were chosen
similarly to what was done for the synthetic dataset. The Kernel bandwidth
$\sigma_1$, ADMM convergence controlling variable $\rho$, fusion penalty
$\gamma$, and $\sigma_2$, all were chosen appropriately after proper
tuning. Kernel Convex Clustering was then applied on the datasets. Using
agglomerative clustering and an elbow plot, we get that the optimal
number of clusters in this case is 7, as shown in
Figure~\ref{fig:elbow_lymphoma}. Originally, there were 9 clusters, but
here we get 7, which does not seem to be a problem, as one or two
clusters had just 3 to 4 points in it. So the clusters were merged to
minimise the entire fit of the data. After getting the number of
clusters, we get the corresponding cluster identities for all points, and
then compute the NMI values for the Lymphoma by comparing the original
and the experimental cluster identities. The NMI value reported in this
case is 0.778. The NMI values for the other methods are given in
Table~\ref{tab:benchmarks}. The comparative study of the t-SNE plots for
the Lymphoma dataset is demonstrated in Figure~\ref{fig:lymphoma_tsne}.

\subsection{Performance on Real Benchmarks}\label{Performance on Real Benchmarks}
To further demonstrate the efficacy of our proposal, we compare KCC with
the peer algorithms on nine benchmark real datasets. The datasets are
taken from the Keel repository \cite{keel} and ASU feature selection
repository \cite{li2018feature}. Preprocessing is first performed on
each dataset before running the KCC algorithm. For datasets with
continuous covariates, we scale the data by centering each of the
datasets and dividing by the corresponding variance. However, for
datasets with categorical variables, centering is not performed, and the
original dataset itself is used. We take the Gaussian kernel, which is a
popular choice for machine learning practitioners. For MNIST, we randomly
select 50 images from 10 classes and apply KCC on the overall 500 data
points. There are four main hyperparameters, $\sigma_1, w_{ij}, \rho,
\gamma$. We tune these four hyperparameters and get the $u_i$'s
corresponding to each point by running KCC. We construct the elbow plots
and get the optimal number of clusters, $K$. Then, using cluster
identities for all points, we compute the NMI values for the dataset by
comparing the original and the experimental cluster identities. The
performance of each algorithm, in terms of NMI, is reported in
Table~\ref{tab:benchmarks}, indicating the superior performance of KCC
on nine real benchmarks.
\section{Conclusion}
In this paper, we designed an algorithm that performs convex clustering in
kernelized Hilbert spaces for datasets in which different groups are
linearly inseparable. KCC utilizes the convexity of the problem to
guarantee convergence to a unique global optimum. More interestingly, we
observe that solving our problem is equivalent to solving the convex
clustering of a finite-dimensional embedding. We try to develop a theory
to characterise the large sample properties of our method, as well as
explore more about the embedding. Our empirical studies on real-life and
synthetic datasets show the success of our method compared to various
standard and recent clustering techniques. This opens many other avenues
for further research. Future directions include multi-kernel extensions, feature weighting for interpretability, and a broader theoretical study correlating infinite- and finite-dimensional embeddings across kernel-based learning frameworks.

\bibliographystyle{apalike}
\bibliography{ref}


\newpage
\appendix

\section*{Appendix}

\section{Problem Statement}
Our objective is to optimize the following expression,
{\small \begin{equation}\label{objective}
\begin{split}
\min_{\bu_1,\dots, \bu_n}\frac{1}{2}\sum_{i=1}^{n}\|\phi(\bx_i)-\bu_i\|^2 + \gamma \sum_{i<j}w_{ij}\|\bu_i - \bu_j\|
\end{split}
\end{equation}}
\section{Necessary Assumptions}
\label{sec_assumption}
We shall apply Hanson-Wright's inequality for Hilbert spaces. We assume that $\bepsilon_i$'s must adopt the following Bernstein's condition so that Hanson-Wright's Inequality is applicable.\\
\\
\textbf{Bernstein's condition on the squared norm:} There exists an universal constant $C>0$ such that
$$\E\left|\|\bepsilon_i\|^2-\E\|\bepsilon_i\|^2\right|^k \le Ck!({L_i\|\Gamma\|_{op}})^{k-2} \|\Sigma_i\|^{2}_{HS}$$
where $\Sigma_i=\E\left[\bepsilon_i\otimes\bepsilon_i\right]$ is the covariance operator of $\bepsilon_i$.
Note that, $\E\left[\langle {\bz}^{\top}\bepsilon,{\bz}^{\top}\bepsilon \rangle\right]=\|\bz\|^2{\sigma}^2$. The matrix $\bA$ that we require in Hanson-Wright's inequality in this case is $\bA={\bz}{\bz}^{\top}$. For any $\delta>0$, let $t=\delta\|\bz\|^2{\sigma}^2$. Now, an easy application of Hanson-Wright's inequality gives us that 
\begin{align*}
    &\mathbb{P}\left[\langle {\bz}^{\top}\bepsilon,{\bz}^{\top}\bepsilon \rangle \ge \E[\langle {\bz}^{\top}\bepsilon,{\bz}^{\top}\bepsilon \rangle]+t\right]\\
    =&\mathbb{P}\left[\langle {\bz}^{\top}\bepsilon,{\bz}^{\top}\bepsilon \rangle \ge(1+\delta)\|\bz\|^2{\sigma}^2 \right]
    \le 2 \exp\left[-C\min\left(\frac{\sigma^4\delta^2}{L^4{\|\Gamma\|}^{2}_{HS}},\frac{\sigma^2\delta}{L^2{\|\Gamma\|}^{2}_{op}}\right)\right]
\end{align*}
In the last inequality, $C>0$ and $L=\max_{1\le i \le n}L_i$, and since $\bepsilon_i$'s are i.i.d., hence $L_i$'s all equal to $L$.

\section{Theoretical Proofs}
\label{kcc_complete_proofs}

\noindent
\textbf{Theorem}~\ref{thm:main}
Let $\phi(\bx_i)=\bu_i+\bepsilon_i$ for all $i=1,\dots,n$, where $\bepsilon_i$ are i.i.d. mean zero sub gaussian random variables in the RKHS $\mathcal{H}$, with respect to the operator $\Gamma$. Let $\hat{\bu^{*}_i}$ be the solutions of \ref{objective}. If $\gamma^{'}\ge \frac{2z_0}{n w_{\min}}$, then 
{\small \[\frac{1}{2n}\sum_{i=1}^{n}\|\hat{\bu^{*}_i}-\bu_i\|^2 \le \frac{3\gamma^{'}}{2}\sum_{i<j}w_{ij}\|\bm{u_i-u_j}\|+{\sigma}^2\left[\frac{1}{n}+\sqrt{\frac{\log(n)}{n^2}}\right]\]}
with probability at least $1-\frac{2}{{n \choose 2}}-2\exp\left[-C\min\left(\frac{\sigma^4\log(n)}{L^4{\|\Gamma\|}^{2}_{HS}},\frac{\sigma^2\sqrt{\log(n)}}{L^2{\|\Gamma\|}^{2}_{op}}\right)\right]$ for some constant $C$.

\begin{proof}

Let $\hat{\bu_i}$ be the estimates of the minimizer of equation \ref{objective}, and let $\bu_i$ be the ground truths. We assume that the projected data points follow the model, $\phi(\bx_i)=\bu_i+\bepsilon_i$, where $\bepsilon_i$ are i.i.d. mean-zero sub-Gaussian random variables in the RKHS $\mathcal{H}$, with respect to the operator $\Gamma$. Additionally, $\E\left[\bepsilon_i\right]=0$, $\E\left[\langle \bepsilon_i \;,\;\bepsilon_i \rangle \right]={\sigma}^2$, and $\E\left[\langle \bepsilon_i \;,\;\bepsilon_j \rangle \right]=0$ for all $i \neq j$.  We define the vectors $\bu=(\bu_1,\dots, \bu_n)^\top$, $\hat{\bu}=(\hat{\bu_1}, \hat{\bu_2},\dots, \hat{\bu_n})^\top, \bm{\phi}=(\phi(\bx_1),\dots,\phi(\bx_n))^\top$ and $\bepsilon=(\bepsilon_1, \dots \bepsilon_n)^\top$. Note that every $\bu_i$ is an element of an RKHS, $\mathcal{H}$. So, we can treat each of them as a function (in the sense of an operator). Hence, $\bu, \hat{\bu}, \bepsilon$ are all $n$ dimensional vectors lying in $\mathcal{H}^n$.  Owing to this notation, we write the following:
{\small \begin{equation}
    \bm{\phi=u+\epsilon}
\end{equation}}
Next, we observe that $\bu_i-\bu_j=(\be_i-\be_j)^\top\bu$ for every pair $i<j$. Let $\bD\in \mathbb{R}^{{n \choose 2}\times n}$ such that $\bD_{ij}=(\be_i-\be_j)^\top $, where $\bD_{ij}$ is the row correspondig to the $(i,j)^{th}$ pair of points. The rows of $\bD$ are spanned by $\be_1-\be_2, \be_2-\be_3,..., \be_{n-1}-\be_n$, which are linearly independent, and thus its rank is $n-1$. Let $\bD=\bU \bSigma \bV_{\beta}^\top$, where $\bU\in \mathbb{R}^{{n \choose 2}\times (n-1)}, \bSigma$ is a $(n-1)\times(n-1)$ diagonal matrix with positive singular values, and $\bV_{\beta} \in \mathbb{R}^{n\times (n-1)}$. Both $\bU$ and $\bV_\beta$ have orthogonal columns. Define $\bV_{\alpha} \in \mathbb{R}^n$, such that $\bV=[\bV_{\alpha} \bV_{\beta}]$ is an orthogonal matrix, i.e. $\bV^{\top}\bV=\bV\bV^{\top}=I$. So ${\bV_{\alpha}}^{\top}\bV_{\beta}=0$ and $\bV_{\alpha}\bV_{\alpha}^\top+\bV_{\beta}\bV_{\beta}^\top=\bI$. We project $\bu$ in the two orthogonal spaces $\bV_{\alpha}$ and $\bV_{\beta}$. Let $\balpha=\bV_{\alpha}^{\top}\bu$ and $\bm{\beta}=\bV_{\beta}^{\top}\bu$. The optimisation now becomes in terms of $\balpha$ and $\bm{\beta}$ as follows:
{\small \begin{equation}\label{eqn 19}
    \|\bm{\phi}-\bV_{\alpha} \balpha-\bV_\beta \bm{\beta}\|^2+\gamma \|\bD\bu\|
\end{equation}}

Recall the matrix $\bD$ defined such that the row of $\bD$ corresponding to the $(i,j)^{th}$ pair is $\bD_{ij}=(\be_i-\be_j)^\top$, where $\be_i$ is the canonical basis element of $\mathbb{R}^{n}$ whose $i^{th}$ entry is 1, and the remaining entries are all 0. Since $\be_1-\be_2, \be_2-\be_3,\dots,\be_{n-1}-\be_n$ span the rows of $D$ and they are linearly independent, so the rank of $\bD$ is $n-1$. Let $\bD=\bU \bSigma {\bV_{\beta}}^{\top}$ be the SVD of $\bD$. $\bU\in \mathbb{R}^{{n \choose 2}\times (n-1)}, \bSigma$ is a $(n-1)\times(n-1)$ diagonal matrix with positive singular values, and $\bV_{\beta} \in \mathbb{R}^{n\times (n-1)}$. Both $\bU$ and $\bV_{\beta}$ have orthogonal columns. Define $\bV_{\alpha} \in \mathbb{R}^n$, such that $\bV=[\bV_{\alpha} \bV_{\beta}]$ is an orthogonal matrix, i.e. $\bV^{\top}\bV=\bV\bV^{\top}=I$. So ${\bV_{\alpha}}^{\top}\bV_{\beta}=0$
We project $\bu$ in the two orthogonal spaces $\bV_{\alpha}$ and $\bV_{\beta}$. Let $\balpha=\bV_{\alpha}^{\top}\bu$ and $\bm{\beta}=\bV_{\beta}^{\top}\bu$. The optimisation now becomes in terms of $\balpha$ and $\bm{\beta}$ as follows:
\begin{equation}\label{eqn 19}
    \|\bm{\phi}-\bV_{\alpha} \balpha-\bV_\beta \beta\|^2+\gamma P(\bm{\beta})
\end{equation}
were $P(\bm{\beta})$ is the fusion penalty.

Note that, $\|\bm{\phi}-\hat{\bu}\|^{2}$ approximates $\|\bm{\phi-u}\|^{2}$, if $\bm{u}$ and $\bm{\hat{u}}$ are close to each other. So the difference between the first two quantities measures the closeness of $\bm{u}$ and $\bm{\hat{u}}$. We see, 
\begin{align*}
    \|\bm{\phi}-\hat{\bu}\|^{2}-\|\bm{\phi}-\bu\|^{2}=&\|\bm{\phi}-\bu+\bu-\hat{\bu}\|^{2}-\|\bm{\phi}-\bu\|^{2}\\=&\|\bm{\phi}-\bu\|^{2}+\|\bm{u-\hat{u}}\|^{2}+2(\bm{\phi-u})^{\bm{T}}(\bm{u-\hat{u}})-\|\bm{\phi-u}\|^{2}\\=&\|\bm{u-\hat{u}}\|^{2}+2 \bm{\epsilon}^{\bm{T}}(\bm{u-\hat{u}})\\=&\|\bm{u-\hat{u}}\|^{2}+2 \bm{\epsilon}^{\bm{T}}\{\bV_{\alpha}(\balpha-\hat{\balpha})+\bV_{\beta}(\bbeta-\hat{\bbeta})\}
\end{align*}

Since $\hat{\bu}$ is the minimiser of our optimization problem. Hence,
\begin{align*}
\|\bm{\phi}-\hat{\bu}\|^2+2\gamma P(\hat{\bu})\le \|\bm{\phi}-\bu\|^2+2\gamma P(\bu)
\implies \|\bm{\phi}-\hat{\bu}\|^2-\|\bm{\phi}-\bu\|^2 \le 2\gamma P(\bu)-2\gamma P(\hat{\bu})
\end{align*}

We have already computed the difference on the left-hand side. Thus,
\begin{align*}
    &\|\bm{u-\hat{u}}\|^{2}+2 \bm{\epsilon}^{\bm{T}}\{\bV_{\alpha}(\balpha-\hat{\balpha})+\bV_{\beta}(\bbeta-\hat{\bbeta})\} \le 2\gamma P(\bu)-2\gamma P(\hat{\bu})\\
    \implies& \frac{\|\bm{u-\hat{u}}\|^{2}}{2n} \le -\frac{\bm{\epsilon}^{\bm{T}}\{\bV_{\alpha}(\balpha-\hat{\balpha})+\bV_{\beta}(\bbeta-\hat{\bbeta})\}}{n}+ \gamma \frac{P(\bu)-P(\hat{\bu})}{n}
\end{align*}

We shall separately bound ${\bepsilon}^{\top}\bV_{\alpha}(\balpha-\hat{\balpha})$ and ${\bepsilon}^{\top}\bV_{\beta}(\bbeta-\hat{\bbeta})$

\textbf{Bounding $\bepsilon^{\top}\bV_{\alpha}(\balpha-\hat{\balpha})$:}
 Note that since $\hat{\balpha}$ and $\hat{\bbeta}$ are the optimal values for our objective, so 
 \begin{align*}
\hat{\balpha}={\bV_{\alpha}}^{\top}(\bm{\phi}-\bV_{\alpha}\hat{\bbeta})
={\bV_{\alpha}}^{\top}(\bV_{\alpha}\balpha+\bV_{\beta}\bbeta+\bepsilon-\bV_{\alpha}\bbeta)=\balpha+{\bV_{\alpha}}^{\top}\bepsilon
\end{align*}
Thus, we get $\bepsilon^{\top}\bV_{\alpha}(\balpha-\hat{\balpha})=\bepsilon^{\top}\bV_{\alpha}{\bV_{\alpha}}^{\top}\bepsilon$. Also, $\E\left[\bepsilon^{\top}\bV_{\alpha}{\bV_{\alpha}}^{\top}\bepsilon\right]=\bsigma^2\|\bV_{\alpha}\|^2=\bsigma^2$ since $\bV_\alpha$ is column of the orthogonal matrix $\bV$. Since $\bV_{\alpha}\bV_{\alpha}^{\top}$ is a symmetric matrix, and $\bepsilon_i$'s are sub-Gaussian in $\mathcal{H}$ and satisfy the assumptions described in Section  \ref{sec_assumption}, we apply Hanson-Wright's inequality on it and get, 
\begin{align*}
&\mathbb{P}\left[\bepsilon^{\top}\bV_{\alpha}{\bV_{\alpha}}^{\top}\bepsilon \ge {\sigma}^2+t \right] 
 \le 2 \exp\left[-C\min\left(\frac{t^2}{L^4{\|\Gamma\|}^{2}_{HS}},\frac{t}{L^2{\|\Gamma\|}^{2}_{op}}\right)\right]
\end{align*}

Note that since $\bV_{\alpha}$ has unit norm, hence $\|\bV_{\alpha}\bV_{\alpha}^{\top}\|_{HS}=\|\bV_{\alpha}\bV_{\alpha}^{\top}\|_{OP}=1$. So there is no term of $\bV_{\alpha}$ in the bound, which generally should occur for Hanson-Wright's inequality.

Now, take $t=\sigma^2\sqrt{\log n}$. Then,
\begin{align*}
\mathbb{P}\left[\frac{\bepsilon^{\top}\bV_{\alpha}{\bV_{\alpha}}^{\top}\bepsilon}{n} \ge {\sigma}^2\left(\frac{1}{n}+\sqrt{\frac{\log n}{n^2}}\right)\right] 
\le 2 \exp\left[-C\min\left(\frac{\sigma^4 \log n}{L^4{\|\Gamma\|}^{2}_{HS}},\frac{\sigma^2 \sqrt{\log n}}{L^2{\|\Gamma\|}^{2}_{op}}\right)\right]
\end{align*}

 \textbf{Bounding ${\bepsilon}^{\top}\bV_{\beta}(\bbeta-\hat{\bbeta}):$}

 Let $\bA=\bU\bSigma$. Note that the columns of $A$ are linearly independent. So its left inverse exists. Let $\bA^{+}$ be the left inverse such that $\bA^{+}\bA=\bI$. Then 
 \begin{align*}
     {\bepsilon}^{\top}\bV_{\beta}(\bbeta-\hat{\bbeta})=&{\bepsilon}^{\top}\bV_{\beta}\bA^{+}\bA(\bbeta-\hat{\bbeta})\\
     =&\sum_{t=1}^{{n \choose 2}}\left\langle{\bepsilon}^{\top}\bV_{\beta}\bA_{*t}^{+},{\bA}_{t*}(\bbeta-\hat{\bbeta})\right\rangle\\
     \le& \sum_{t=1}^{{n \choose 2}}\|{\bepsilon}^{\top}\bV_{\beta}\bA_{*t}^{+}\|\|{\bA}_{t*}(\bbeta-\hat{\bbeta}\|\\
     \le & \left\{\max_{t=1,\dots,{n \choose 2}}\|{\bepsilon}^{\top}\bV_{\beta}\bA_{*t}^{+}\|\right\}\left\{\sum_{t=1}^{{n \choose 2}}\|{\bA}_{t*}(\bbeta-\hat{\bbeta}\|\right\}
 \end{align*}
Let $\ba_{t}=\bV_{\beta}\bA_{*t}^{+}$. Take $\bz=\ba_t$ and applying Hanson-Wright's inequality described in Section \ref{sec_assumption} for any $\delta>0$ we get,

\begin{align*}
\mathbb{P}\left[\bepsilon^{\top}\ba_{t}{\ba_{t}}^{\top}\bepsilon \ge (1+\delta){\sigma}^2\|\ba_t\|^2 \right] \le 2 \exp\left[-C\min\left(\frac{\sigma^4\delta^2}{L^4{\|\Gamma\|}^{2}_{HS}},\frac{\sigma^2\delta}{L^2{\|\Gamma\|}^{2}_{op}}\right)\right]
\end{align*}

Note that, $\ba_t \in \mathbb{R}^{n}$. 

Also, the above holds for $\forall \delta>0$. Now, choose $\delta_0$ such that $\exp\left[-C\min\left(\frac{\sigma^4\delta_{0}^2}{L^4{||\Gamma||_{HS}}^2},\frac{\sigma^2\delta_0}{L^2{||\Gamma||_{op}}^2}\right)\right]=\frac{1}{{n \choose 2}^2}$. It is easy to see that $\delta_{0}>0$. So,
 \begin{align*}
 \mathbb{P}\left[\bepsilon^{\top}\ba_t{\ba_t}^{\top}\bepsilon \ge (1+\delta_0){\sigma}^2\|\ba_t\|^2 \right] \le \frac{2}{{n \choose 2}^2}
 \end{align*}
Let $z^{2}_{0}=\max_{t=1,\dots,{n \choose 2}}(1+\delta_0)\sigma^2\|\ba_t\|^2$. Then, for any $t \in \{1,\dots,{n \choose 2}\}$, 
$$\bepsilon^{\top}\ba_t{\ba_t}^{\top}\bepsilon \ge z^{2}_{0}\ge (1+\delta_0){\sigma}^2\|\ba_t\|^2$$ 
and hence,
$$\mathbb{P}\left[\bepsilon^{\top}\ba_t{\ba_t}^{\top}\bepsilon \ge z^{2}_0\right]\le \mathbb{P}\left[\bepsilon^{\top}\ba_t{\ba_t}^{\top}\bepsilon \ge (1+\delta_0){\sigma}^2\|\ba_t\|^2 \right] \le \frac{2}{{n \choose 2}^2}$$   
Also, by union bound
\begin{align*}
\mathbb{P}\left[\max_{t=1,\dots,{n \choose 2}}\|{\ba_t}^{\top}\bepsilon\|^2 \ge z^{2}_0\right] \le \sum_{t=1}^{{n \choose 2}}\mathbb{P}\left[\|{\ba_t}^{\top}\bepsilon\|^2 \ge z^{2}_0\right] \le \frac{2}{{n \choose 2}}
\end{align*}
If $\gamma^{'}\ge \frac{2z_0}{n w_{\min}}$, then 
\begin{align*}
    \mathbb{P}\left[\max_{t=1,\dots,{n \choose 2}}\frac{1}{n}\|\bepsilon^\top \ba_t\|\ge \frac{w_{\min}\gamma^{'}}{2}\right] \le \mathbb{P}\left[\max_{t=1,\dots,{n \choose 2}}\frac{1}{n}\|\bepsilon^\top \ba_t\|\ge \frac{z_0}{n}\right] \le  \frac{2}{{n \choose 2}}
\end{align*}
Thus, $\max_{t=1,\dots,{n \choose 2}}\frac{\|\bepsilon^\top \ba_t\|}{n}\ge \frac{w_{\min}\gamma^{'}}{2}$ with probability at max $\frac{2}{{n \choose 2}}$

\noindent
Thus, we have,
\begin{align*}
\frac{\|\bu-\hat{\bu}\|^2}{2n}\le & \frac{1}{n}\bepsilon^\top \bV_{\alpha}\bV^{\top}_{\alpha}\bepsilon+\frac{1}{n}\bepsilon^{\top}\bV_{\beta}\left(\hat{\bbeta}-\bbeta\right)+\gamma^{'}\left[P(\bu)-P(\hat{\bu})\right] \\ \le & {\sigma}^2\left(\frac{1}{n}+\sqrt{\frac{\log n}{n^2}}\right)+\frac{\gamma^{'}w_{\min}}{2}\sum_{t=1}^{{n \choose 2}}\|\bA_{t*}(\bbeta-\hat{\bbeta})\|
\end{align*}
with probability at least $1-\frac{2}{{n \choose 2}}-2 \exp\left[-C\min\left(\frac{\sigma^4 \log n}{L^4{\|\Gamma\|}^{2}_{HS}},\frac{\sigma^2 \sqrt{\log n}}{L^2{\|\Gamma\|}^{2}_{op}}\right)\right]$

We finally use the fact that $w_{\min}<w_{ij}$ for all pairs $i,j$ to get the $w_{ij}$ terms inside the summation. That is,
$$\frac{\gamma^{'}w_{\min}}{2}\sum_{t=1}^{{n \choose 2}}\|\bA_{t*}(\bbeta-\hat{\bbeta})\|\le \frac{\gamma^{'}}{2}\sum_{i<j}w_{ij}\|\bA_{ij*}(\bbeta-\hat{\bbeta})\|$$
The triangle inequality can finally be employed to get the final result as mentioned in the main text.
\end{proof}




\section{Sensitivity Analysis of Hyperparameters for Synthetic Dataset}
\label{sec:app_hyp_sens}
We provide the details of the sensitivity analysis of the synthetic dataset in Figure \ref{fig:hyper_sens}. We experimented on a large range of values for these 4 hyperparameters, constructed an elbow plot in each case to get the number of clusters, and finally tried to see how they affect the number of
clusters. This is illustrated below, in Figure 1. To check the variation with respect to a particular hyperparameter, say $\sigma_1$, we select various other triplets corresponding to $(\sigma_2, \rho, \gamma)$; now for each
such triplet we vary $\sigma_1$, get the centroids, construct the elbow plots
and finally the number of clusters, which turns out to be 5. This
process is repeated for all three remaining hyperparameters.
The number of clusters consistently comes out to be 5 in all four
cases. We tune all these 4 hyperparameters, and get the optimal values
of $\sigma_1 = 1, \sigma_2 = 100, \gamma = 1, \rho = 0.001$. 

\section{Effect of Different Kernels on KCC}
\label{sec:app_diff_ker}
We conducted additional experiments using a polynomial kernel of the form 
$k(\bx,\bx^{'})=(c+x^\top x^{'})^d$. We varied the degree $d$ of the polynomial to be within the range $\{2,3,4,5,6\}$, and tuned the bias term $c$ within the range $\{-1,-0.95,-0.9,\dots,0.9,0.95,1\}$, selecting the best configuration based on NMI. The results, summarized in the Table~\ref{tab:app_ker_change}, show that the polynomial kernel achieves performance comparable to the Gaussian kernel on several datasets, while exhibiting some variability depending on the data characteristics.
    
These results indicate that the method is not restricted to the Gaussian kernel and extends naturally to other non-linear kernels. 
\begin{table}[!ht]
\label{Kernel_choice}
\centering
\scriptsize
\caption{Performance analysis of five datasets for three different kernels.}
\label{tab:app_ker_change}
\begin{tabular}{l|cccccccc}
\toprule
Ker. / Data & Lymphoma & OrlRaws10P & Lung & GLIOMA & MNIST\\
\midrule
Gaussian &0.78 & 0.85 & 0.80 &  0.53 & 0.61 \\
Polynomial & 0.85 & 0.80 & 0.61 & 0.45 & 0.54 \\
Hyperbolic & 0.81 & 0.76 & 0.78 & 0.41 & 0.57 \\
\bottomrule
\end{tabular}
\end{table}

\begin{figure}[!ht]
\centering
\includegraphics[width=0.50\linewidth]{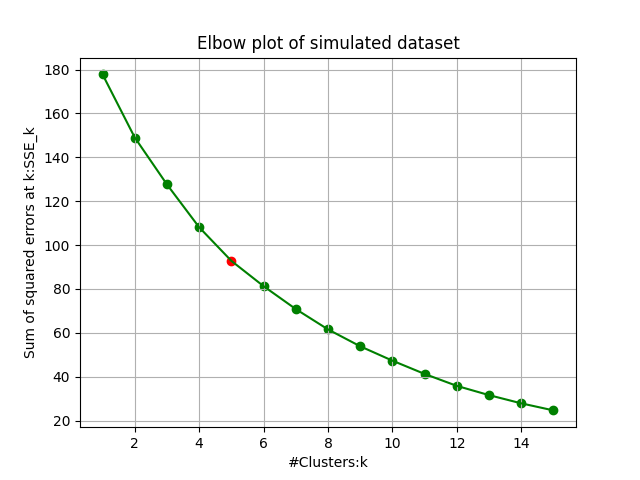}
\caption{Elbow plot of synthetic dataset with $\sigma_1=1, \sigma_2=100, \gamma=1, \rho=0.001$. This set of values gives the optimal clustering with NMI of 1.}
\label{fig:elbow_synthetic}
\end{figure}

\begin{figure}[!ht]
    \centering
    \begin{subfigure}[b]{0.23\textwidth}\label{orig}
        \centering
        \includegraphics[width=\textwidth]{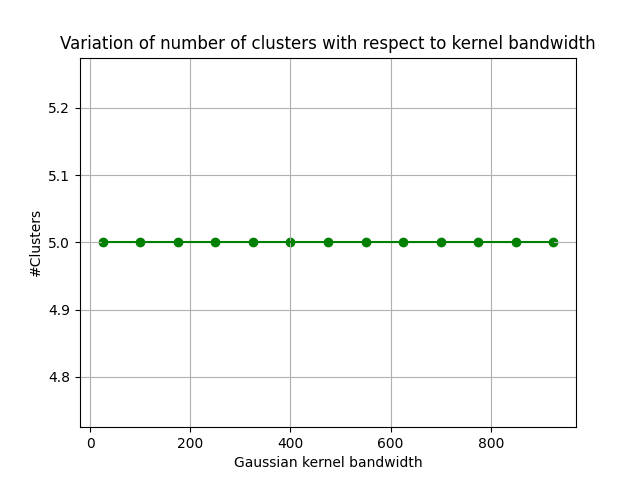}
        \caption{Variation of number of clusters with changing $\sigma_1$ keeping others fixed}
    \end{subfigure}
    \hfill
    \begin{subfigure}[b]{0.23\textwidth}\label{kmeans}
        \centering
        \includegraphics[width=\textwidth]{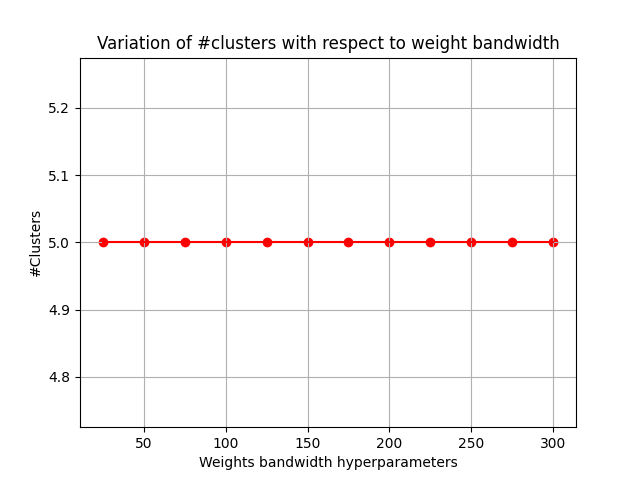}
        \caption{Variation of number of clusters with changing $\sigma_2$ keeping others fixed}
    \end{subfigure}
    \hfill
    \begin{subfigure}[b]{0.23\textwidth}
        \centering
        \includegraphics[width=\textwidth]{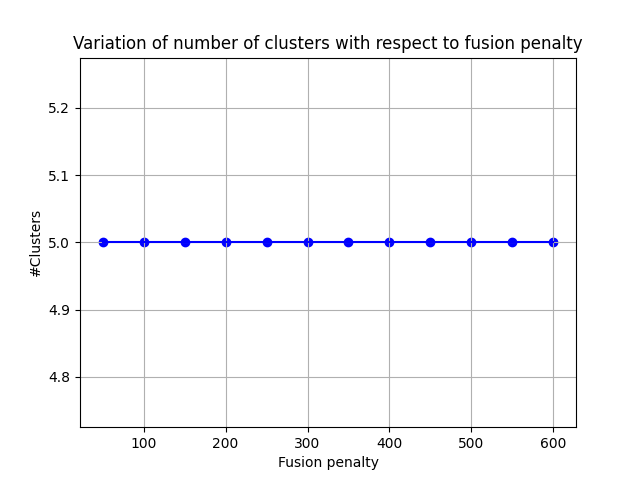}
        \caption{Variation of number of clusters with changing $\gamma$ keeping others fixed}
    \end{subfigure}
    \hfill
    \begin{subfigure}[b]{0.23\textwidth}\label{KCC}
        \centering
        \includegraphics[width=\textwidth]{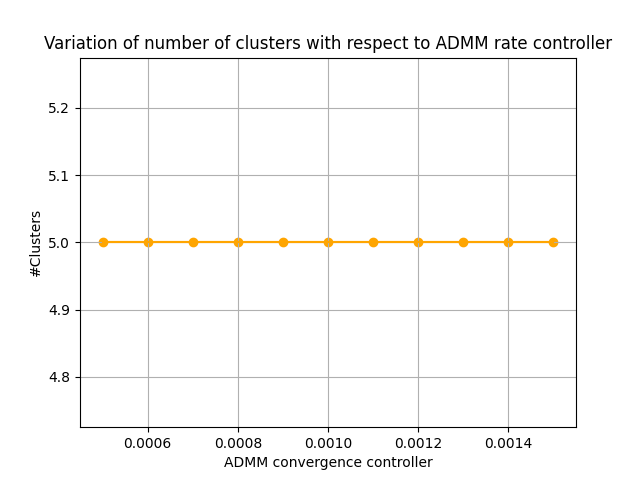}
        \caption{Variation of number of clusters with changing $\rho$ keeping others fixed}
    \end{subfigure}
    \caption{\small{Variation of the number of clusters with each individual hyperparameter fixing others. For checking the dependence with respect to a hyperparameter, various triplets corresponding to the remaining hyperparameters were chosen; then for each triplet, the main hyperparameter was varied over a long range of values, of which we have illustrated just a few. The total number of clusters remains 5 across all four separate experiments.}}
    \label{fig:hyper_sens}
\end{figure}

\section{Remarks on Future Works}
\paragraph{Incorporation of Feature Selection in KCC.}
In order to introduce feature selection in KCC, we introduce the concept of feature weighting, where the $j^{th}$ feature has weight $w_j$. Thus, the kernel map $\phi_{\bw}$ highlights the dependence of $\bw$. Further, we shall select $\phi_{\bw}$ so that the associated kernel follows $k(\bx,\by)=f(\|\bx-\by\|_{\bw})$, for some continuously differentiable function $f$. This assumption will help us in optimizing the objective with respect to $\bw$ by suitably differentiating it. We shall devise a coordinate descent algorithm to optimize the final objective, where in one step we fix $\bw$ and apply the standard KCC algorithm to find the centroids $\bu_i$'s. In the other step, we fix the $\bu_i$'s and estimate $\bw$ under the constraint that $\bm{1}^\top \bw=1$. This can be done by constructing a majorizer of the objective by leveraging the concavity of the penalty term and minimizing it.

\paragraph{Scalability of KCC.}
While this work focuses on the theoretical and empirical validation of KCC, that scalability is vital for large datasets, and standard kernelized convex clustering suffers significantly in situations where the dataset is of a large size. This is mainly due to the costly eigen-decomposition of the kernel matrix, $\bK$, or finding its inverse, ${\bK}^{-1}$. KCC's framework can incorporate low-rank approximations, such as the \textit{Nyström} method or random Fourier features, to reduce complexity from $\mathcal{O}(n^2)$ to $\mathcal{O}(nm)$, where $m \ll n$. This is compatible with the ADMM-based optimization, as the embedding $\mathbf{Z}$ can be computed via approximate factorization. Large-scale applications thus turn out to be a potential direction for future work.

\paragraph{Theoretical guarantees of cluster recovery.}
The main focus of this work was mainly on the theoretical and empirical validation of KCC, we have established an explicit equivalence between KCC and convex clustering, for which theoretical guarantees of cluster recocery have already been extensively studied. This connection suggests that analogous guarantees for our framework could likely be obtained by adapting existing results from the convex clustering literature. Making these guarantees explicit is an important and interesting direction for future work.

\section{System Configuration}
We performed all experiments on an NVIDIA RTX GeForce 3090 GPU with $24$ GB of memory and $64$ GB of RAM.

\clearpage

\end{document}